\pdfoutput=1

\documentclass[11pt]{article}

\usepackage[preprint]{acl}

\usepackage{times}
\usepackage{latexsym}

\usepackage[T1]{fontenc}

\usepackage[utf8]{inputenc}

\usepackage{microtype}

\usepackage{inconsolata}
\usepackage{graphicx}
\usepackage{booktabs}
\usepackage{multirow}
\usepackage{amsmath}
\usepackage{soul}
\soulregister\cite7
\soulregister\ref7 
\soulregister\citep7 
\soulregister\citet7 
\soulregister\pageref7 
\usepackage{colortbl}  
\usepackage{array}

\usepackage{graphicx}
\usepackage{enumitem}
\usepackage{pifont}
\usepackage{algorithm}
\usepackage[noend]{algpseudocode}
\usepackage{amsthm}
\usepackage{ifthen}

\usepackage{amsthm}
\usepackage{amssymb}

\newcommand{\mytextcolor}[1]{\textcolor{black}{#1}}

\usepackage{arydshln}

\title{\textsc{P-MMEval}: A Parallel Multilingual Multitask Benchmark\\for Consistent Evaluation of LLMs}

\author{Yidan Zhang\thanks{~~Work was done when Yidan Zhang and Boyi Deng were interning at Tongyi Lab, Alibaba Group Inc. Corresponding author: Yu Wan.}~~~~~Yu Wan$^{*}$~~~~~Boyi Deng$^{*}$~~~~~Baosong Yang~~~~~Haoran Wei~~~~~Fei Huang$^{a}$\thanks{~~Google Scholar IDs of Fei Huang$^{a}$ and Fei Huang$^{b}$ are \href{https://scholar.google.com/citations?user=7udAEzMAAAAJ}{7udAEzMAAAAJ} and \href{https://scholar.google.com/citations?user=9r98PpoAAAAJ}{9r98PpoAAAAJ}, respectively.}\\
\textbf{Bowen Yu}~~~~~\textbf{Junyang Lin}~~~~~\textbf{Fei Huang}$^{b\dagger}$~~~~~\textbf{Jingren Zhou} \\
    Tongyi Lab, Alibaba Group Inc \\
    \texttt{\{nianjun.zyd,wanyu.wy,dengboyi.dby\}@alibaba-inc.com}
}

\begin{document}
\maketitle
\begin{abstract}
Recent advancements in large language models (LLMs) showcase varied multilingual capabilities across tasks like translation, code generation, and reasoning. Previous assessments often limited their scope to fundamental natural language processing (NLP) or isolated capability-specific tasks. To alleviate this drawback, we aim to present a comprehensive multilingual multitask benchmark. First, we introduce \textsc{P-MMEval}, a large-scale benchmark covering effective fundamental and capability-specialized datasets. Furthermore, \textsc{P-MMEval} delivers consistent language coverage across various datasets and provides parallel samples.
Finally, we conduct extensive experiments on representative multilingual model series to compare performances across models and tasks, explore the relationship between multilingual performances and factors such as tasks, model sizes, languages, and prompts, and examine the effectiveness of knowledge transfer from English to other languages. The resulting insights are intended to offer valuable guidance for future research. The dataset is available at \href{https://huggingface.co/datasets/Qwen/P-MMEval}{https://huggingface.co/datasets/Qwen/P-MMEval}

\end{abstract}

\section{Introduction}
\mytextcolor{In recent years, large language models~\citep[LLMs,][]{gpt3,gpt4,llama2,cloud,qwen} have raised significant interest in the artificial intelligence (AI) community.
As most LLMs are English-centric, when we focus on the performances of a specific LLM, it generally refers to the evaluation results on English benchmarks.
For example, early research focuses on reporting evaluation results on fundamental natural language processing (NLP) benchmarks. i.e, how accurately the LLM understands and generates text, including \textsc{TriviaQA} \cite{triviqa}, \textsc{WinoGrande} \cite{winogrande}, and \textsc{HellaSwag} \cite{hellaswag}.
Nowadays, researchers are more interested in capability-specialized benchmarks, i.e., how well LLM performs on a group of specific task-solving problems, including \textsc{GSM8K} \cite{gsm8k} for mathematical reasoning, \textsc{MMLU} \cite{mmlu} for knowledge acquisition, and \textsc{HumanEval} \cite{humaneval} for code generation.
However, there is currently little work on \mytextcolor{systematically} evaluating the multilingual capabilities of LLMs.
When developing and iterating LLMs, giving accurate and parallel evaluation results is crucial for identifying their multilingual capabilities and quantifying their performance.}

\mytextcolor{Building a benchmark with both inclusive task coverage and strong linguistic parallelism is difficult.
Measuring the multilingual abilities of a specific LLM, or comparing the quality of generated multilingual responses from one LLM to another, remains a big challenge in developing multilingual LLMs.
Early work focuses on an isolated evaluation pipeline for a specific task, or to be more concrete, a specific perspective of LLM abilities: \textsc{MHellaSwag} \cite{eval-harness} aims at collecting the multilingual understanding abilities, \textsc{XLSum} \cite{XLSum} mainly focus on evaluating the quality of generated multilingual text, \textsc{HumanEval-XL} \cite{humaneval-xl} is used for quantify how well-executed the generated code segments are, and \textsc{MGSM} \cite{mgsm} is made for testifying the performance on arithmetic reasoning.
In modern research, for delivering simpler aggregation and comprehensive evaluation when judging model abilities, researchers collect several popular isolated benchmark tasks and propose a united, large-scale multilingual benchmark system like XTREME \cite{xtream}, XTREME-R \cite{xtream-r}, XGLUE \cite{xglue}, MEGA \cite{mega}, and BUFFET \cite{buffet} for multi-task assessments.
However, these large-scale benchmarks 1) are tailored predominantly to fundamental NLP tasks and 2) inconsistently cover multiple languages across their selected datasets. }

\mytextcolor{In this paper, our goal is to develop a comprehensive multilingual multitask benchmark.
To this end, we first include three datasets from fundamental NLP tasks covering both understanding and generation. The second phase of our endeavor involves a meticulous curation of the most intensely studied capability-specialized tasks in contemporary research including code generation, knowledge comprehension, mathematical reasoning, logical reasoning, and instruction following.
Finally, we construct a collection of datasets \textsc{P-MMEval}, consisting of three fundamental NLP datasets and five advanced capability-specialized datasets.
To maintain language coverage among all selected datasets, we unify 10 languages considering the cost and computational limitations via expert translation review to construct the missing multilingual portions.}

\mytextcolor{To summarize, our contributions are as follows:
\begin{itemize}
    \item We develop a multilingual multi-task benchmark \textsc{P-MMEval} that includes both fundamental and capability-specialized tasks, which ensures consistent language coverage across various datasets and provides parallel samples across different languages. This benchmark facilitates a thorough assessment of multilingual capabilities and enables unprecedented fairness and consistency in evaluating cross-lingual transfer capabilities.
    \item Our experiments offer a comprehensive analysis of the multilingual capabilities of various LLMs, showcasing performance across different prompts, models, languages, and tasks. Our analyses underscore a significant benchmark sensitivity in evaluating multilingual capabilities, indicating that the ``nativeness" of the benchmark dramatically affects the observed multilingual evaluation results. 
    \item We introduce the cross-lingual accuracy consistency ratio (CACR) to analyze the effectiveness of knowledge transfer from English to other languages across various target languages and task scenarios. Our analysis indicates that, among the tested tasks, code knowledge is the easiest to transfer, while logical reasoning proves the most difficult. Regarding specific languages, transfer is facilitated by linguistic similarity. 
\end{itemize}}

\begin{table*}[t]
\centering
\scalebox{0.8}{
    \begin{tabular}{llllll}
        \toprule
            \mytextcolor{\textbf{Source}} & \mytextcolor{\textbf{Task}}                                     & \textbf{Benchmarks}      & \mytextcolor{\textbf{\# Examples}} &  Test sets & Metric \\
        \midrule
            \multirow{1}{*}{\mytextcolor{Existing}}           & \multirow{1}{*}{Generation}              
                                                     & \textsc{Flores-200}~\cite{flores200}   & 1012 $\times$ 10  & Annotation                                         & BLEU   \\
        \midrule
            \multirow{7}{*}{\mytextcolor{Extension}}            & \multirow{2}{*}{Understanding}           & \mytextcolor{XNLI}~\cite{xnli}             & 120 $\times$ 10 (3)   & Translation                                         & Acc    \\
                                                    & & \mytextcolor{\textsc{MHellaSwag}}~\cite{eval-harness}  & 120 $\times$ 10 (3)   & Translation                                         & Acc    \\
        \cmidrule{2-6}
            & Code generation                          & \textsc{HumanEval-XL}~\cite{humaneval-xl} & 80 $\times$ 10 (3) $\times$ 12  & Translation                                         & Pass@1 \\
        \cmidrule{2-6}
            & Mathematical reasoning                   & \textsc{MGSM}~\cite{mgsm}         & 250 $\times$ 10 (3)   & Translation                                         & Acc    \\
        \cmidrule{2-6}
            & Logic reasoning                          & \textsc{MLogiQA}~\cite{logiqa}       & 80 $\times$ 10 (8)   & Translation                                         & Acc    \\
        \cmidrule{2-6}
            & \mytextcolor{Knowledge} & \textsc{MMMLU}~\cite{mmlu}   & 400 $\times$ 10 (2)  & Translation                                         & Acc    \\

        \cmidrule{2-6}
            & \mytextcolor{Instruction following} & \textsc{MIFEval}~\cite{ifeval}   & 96 $\times$ 10 (9)  & Translation                                         & Acc    \\
        \bottomrule
    \end{tabular}
}

\caption{\mytextcolor{An overview of the \textsc{P-MMEval} benchmark. In total, \textsc{P-MMEval} takes seven multilingual tasks into consideration, which is built on eight benchmarks. ``\# Examples'' denotes ``the number of examples per language'' $\times$ ``the number of involved languages'' $\times$ ``the number of programming languages'' (special for \textsc{HumanEval-XL}), and the numbers of extended languages are in parentheses. \mytextcolor{``}Test sets\mytextcolor{''} section describes the nature of the test sets (whether they are translations of English data or independently annotated).}}
\label{tab:benchmark}
\end{table*}

\section{Related Work}
\paragraph{Isolated Fundamental NLP Benchmarks}
\mytextcolor{Although diverse multilingual evaluation benchmarks have been established, they focused on basic language understanding and generation capabilities of models. Notable work includes \textsc{XNLI} \cite{xnli} for natural language inference, \textsc{XCOPA} \cite{xcopa}, \textsc{MHellaSwag} \cite{eval-harness}, and \textsc{XWinograd} \cite{xwinograd} for commonsense reasoning, \textsc{PAWS-X} \cite{paws-x} for paraphrase identification, \textsc{XL-WiC} \cite{xl-wic} for word sense disambiguation, as well as the span extraction QA datasets including \textsc{XQUAD} \cite{xquad}, \textsc{MLQA} \cite{mlqa}, and \textsc{TyDiQA-GoldP} \cite{tydiqa}. Additional examples include \textsc{XLSum} \cite{XLSum} for text summarization and \textsc{Flores-200} \cite{flores200} for machine translation. Each of those benchmarks is typically designed for a specific task, solely focusing on one aspect of the model's capabilities.}

\paragraph{Unified Fundamental NLP Benchmarks}
\mytextcolor{There are also large-scale benchmarks that unify diverse existing datasets, aiming at offering a comprehensive evaluation of the model's abilities from various perspectives. For instance, XTREME \cite{xtream} comprises four tasks related to natural language understanding (NLU). Its refined version, XTREME-R \cite{xtream-r}, optimizes the specific datasets tailored for each task category within XTREME. The XGLUE \cite{xglue}, MEGA \cite{mega}, and BUFFET \cite{buffet} benchmarks integrate various datasets for both understanding and generation tasks.}

\paragraph{Capability-specialized Multilingual Benchmarks}
\mytextcolor{The advanced task-solving capabilities of LLMs have garnered significant attention from the research community. The six capabilities that receive the most emphasis are mathematical reasoning \cite{gsm8k, math}, logical reasoning \cite{logiqa}, instruction following \cite{alpaca_eval}, knowledge comprehension \cite{mmlu}, code generation \cite{humaneval}, and conversational abilities \cite{mtbench}. Typical multilingual benchmarks include MGSM \cite{mgsm} for mathematical reasoning, the OpenAI multilingual version of MMLU (MMMLU)\footnote{\href{https://huggingface.co/datasets/openai/MMMLU}{https://huggingface.co/datasets/openai/MMMLU}} for knowledge comprehension, and \textsc{HumanEval-XL} \cite{humaneval} for code generation. }

\mytextcolor{All the benchmarks mentioned above focus either exclusively on fundamental NLP capabilities or on advanced application abilities. Additionally, there is inconsistent multilingual coverage across various datasets within a single multi-task benchmark. The proposed benchmark \textsc{P-MMEval} integrates three fundamental NLP datasets and five capability-specialized datasets, providing consistent language coverage across all selected datasets.}

\section{P-MMEval}
\label{benchmark}
\mytextcolor{We aim to build a comprehensive evaluation system that unifies diverse NLP and capability-specialized tasks, ensures consistent language coverage per task, and offers parallel samples across languages to facilitate consistent comparisons. The overview of our proposed \textsc{P-MMEval} is shown in Table \ref{tab:benchmark}. }

\subsection{Design Principles}
\label{design_principles}

\textbf{Diversity in tasks}
First, the two key fundamental NLP tasks of generating and understanding are covered. More critically, through in-depth analysis, we identify and establish five kinds of core capabilities of current LLMs, including code generation, knowledge comprehension, mathematical reasoning, logical reasoning, and instruction following. 

\textbf{Diversity in languages}
\mytextcolor{To ensure that our benchmark can also help testify the cross-lingual transferability of LLMs, we unify 10 different languages spanning 7 language families, including English (\textit{en}), Chinese (\textit{zh}), Arabic (\textit{ar}), Spanish (\textit{es}), Japanese (\textit{ja}), Korean (\textit{ko}), Thai (\textit{th}), French (\textit{fr}), Portuguese (\textit{pt}), and Vietnamese (\textit{vi}). }

\subsection{Fundamental NLP Dataset Curation}
In light of the diversity of fundamental NLP datasets, we meticulously select three datasets widely employed in research \cite{mega,buffet,xglue}, spanning across the two major categories of understanding and generation. Below, we briefly summarize these three datasets.

i) \textsc{XNLI}: The natural language inference (NLI) dataset, \textsc{XNLI} \cite{xnli}, involves classifying whether a hypothesis is entailed, contradicted, or unrelated to the premise.

ii) \textsc{MHellaSwag}: The commonsense reasoning dataset \textsc{MHellaSwag} \cite{hellaswag} consists of sentences or paragraphs, requiring models to predict the most likely option to complete the sentence or paragraph ending.

iii) \textsc{Flores200}: The multilingual machine translation \textsc{Flores200} \cite{flores200} is an evaluation benchmark for low-resource and multilingual machine translation.

\begin{table*}[t]
\centering
{\setlength{\tabcolsep}{5mm}
\scalebox{0.65}{%
\begin{tabular}{ccccccccccc}
\toprule
Dataset      & \textit{zh}    & \textit{ar}    & \textit{es}    & \textit{ja}    & \textit{ko}    & \textit{th}    & \textit{fr}    & \textit{pt}    & \textit{vi} \\ \midrule
XNLI         & -     & -     & -     & 22.50 & 11.67 & -     & -     & 10.83 & -   \\ 
\textsc{MHellaSwag}   & -     & -     & -     & 82.50 & 77.50 & 26.67 & -     & -     & -  \\ 
\textsc{HumanEval-XL} & -     & -     & -     & 42.50 & 23.75 & 31.25 & -     & -     & -  \\ 
MGSM         & -     & ~~9.20  & -     & -     & 32.80 & -     & -     & ~~5.60  & 27.20 \\ 
\textsc{MLogiQA}      & -     & 22.50 & 30.00  & 51.25 & 33.75 & 46.25 & ~~3.75  & 46.25 & 18.75 \\ 
MMMLU        & -     & -     & -     & -     & -     & 26.00 & 13.50 & -     & -  \\
\textsc{MIFEval}      & 25.50 & 23.81 & 20.00 & 45.71 & 36.19 & 37.14 & 21.90 & 17.14 & 24.76\\ \bottomrule
\end{tabular}
}
}

\caption{The table presents the percentage of modifications made by professional translators to the machine translation results. The symbol ``-'' indicates that there are samples in the corresponding language and no translation construction is required.}
\label{tab:expert_review}
\end{table*}

\subsection{Capability-specialized Dataset Curation}
\mytextcolor{Besides the fundamental NLP tasks mentioned above, we also select one dataset for each of the five capability-specialized tasks. In detail, the involved specialized capabilities in \textsc{P-MMEval} are:}

\begin{itemize}
    \item \textbf{Code generation} We utilize \textsc{HumanEval-XL} \cite{humaneval-xl} dataset, which establishes connections between 23 natural languages (NLs) and 12 programming languages (PLs). 
    \item \textbf{Mathematical reasoning} We use the \textsc{MGSM} \cite{mgsm} dataset, a multilingual version translated from the monolingual \textsc{GSM8k} dataset consisting of math word problems. 
    \item \textbf{Logical reasoning} We keep the original English and Chinese examples from origin \textsc{LogiQA} \cite{logiqa} dataset. 
    \item \textbf{Knowledge aqcuisition} We sample a subset of MMMLU comprising 200 ``hard'' samples and 200 ``easy'' samples. The performance of six diverse models (\textsc{Qwen2.5-7B, Qwen2.5-72B, LLaMA3.1-8B, LLaMA3.1-70B, Mistral-Nemo}, and \textsc{Mistral-Large}) is utilized as a proxy for selecting ``hard'' and ``easy'' samples. Concretely, we compile an ``easy'' subset comprising 6,335 instances where all models excel, and a ``hard'' subset consisting of 663 instances that challenge every model. Subsequently, guided by annotations from \textsc{MMLU-Redux} \cite{mmlu-redux}, we refine these subsets by discarding 798 erroneous instances from the ``easy'' pool and 160 from the ``hard'' pool. Finally, we systematically sample 200 instances from each of the pruned pools, thus creating our finalized ``easy'' and ``hard'' evaluation sets. 
    \item \textbf{Instruction following} We employ the English \textsc{IFEval} \cite{logiqa} dataset, which consists examples following pre-defined 25 types of ``verifiable instruction''. 
\end{itemize}

\subsection{Expansion of the Selected Datasets}
To maintain consistency across all languages, we extend the support of some benchmark datasets on the missing languages by collecting human-annotated translation results. The number of expanded languages and samples for each dataset is listed in the ``\#Example'' column of Table \ref{tab:benchmark}. More details of sampling are provided in Appendix Section \ref{data_sample_process}.  

We initially generate translated examples using the advanced \textsc{GPT-4o}\footnote{\texttt{gpt-4o-2024-05-13}} model. Subsequently, a professional translation team conducts an exhaustive review of the machine translation outputs, correcting any errors, localizing vocabulary, and removing instances that do not translate well across languages. This meticulous process ensures both high translation quality and cultural adaptability.

The modification rate by post-review is detailed in Table \ref{tab:expert_review}. It is apparent that datasets contain translation errors to varying extents, with error rates peaking at 82.50\%. This underscores the limitations of using raw machine-generated translations for dataset extension, highlighting the critical need for human review to maintain translation fidelity. Notably, among the most frequent errors are mistranslations of proper nouns and inconsistencies in terminology usage, followed by omissions. These trends indicate that the model currently struggles with specific domain terminology and maintaining contextual coherence.

\begin{table*}[t]
\centering
\scalebox{0.6}{%
\begin{tabular}{ccccccccccc}
\toprule
\multirow{2}{*}{Model} & \multicolumn{2}{c}{Understanding} & \begin{tabular}[c]{@{}c@{}}Code \\ generation\end{tabular} & \begin{tabular}[c]{@{}c@{}}Mathematical \\ reasoning\end{tabular} & \begin{tabular}[c]{@{}c@{}}Logic \\ reasoning\end{tabular} & Knowledge & \begin{tabular}[c]{@{}c@{}}Instruction \\ following\end{tabular} & Generation & \multirow{2}{*}{AVG\_S} & \multirow{2}{*}{AVG\_U} \\ 

\cmidrule(l{3pt}r{3pt}){2-3}
\cmidrule(l{3pt}r{3pt}){4-4}
\cmidrule(l{3pt}r{3pt}){5-5}
\cmidrule(l{3pt}r{3pt}){6-6}
\cmidrule(l{3pt}r{3pt}){7-7}
\cmidrule(l{3pt}r{3pt}){8-8}
\cmidrule(l{3pt}r{3pt}){9-9}

                       & \textsc{XNLI}          & \textsc{MHellaSwag}        & \textsc{HumanEval-XL}                                               & \textsc{MGSM}                                                              & \textsc{MLogiQA}                                                    & \textsc{MMMLU}     & \textsc{MIFEval}                                                          & \textsc{Flores-200} &                         &                         \\ \midrule

\multicolumn{11}{c}{\textit{Open-source models (\textless{}7B)}} \\
\cdashline{1-11}\noalign{\vskip 0.3ex}
\textsc{LLaMA3.2-1B}            & 31.67         & 24.49             & 37.71                                                      & 12.08                                                             & 27.12                                                      & 27.80     & 35.42                                                            & 29.30      & 28.03                   & 28.08                   \\ 
\textsc{LLaMA3.2-3B}            & 30.67         & 23.74             & 37.42                                                      & 11.64                                                             & 25.62                                                      & 26.85     & 34.90                                                            & \textbf{36.85}      & 27.29                   & 27.21                   \\ 
\textsc{Qwen2.5-0.5B}           & 22.25         & 19.68             & 33.92                                                      & 13.12                                                             & 14.62                                                      & 30.25     & 30.21                                                            & 15.95      & 24.42                   & 20.97                   \\ 
\textsc{Qwen2.5-1.5B}           & 46.58         & 36.35             & 48.59                                                      & 35.20                                                             & 35.12                                                      & 42.02     & 44.37                                                            & 21.37      & 41.06                   & 41.47                   \\ 
\textsc{Qwen2.5-3B}             & 60.08         & 48.09             & 60.75                                                      & 69.40                                                             & 39.38                                                      & 46.27     & 66.46                                                            & 25.75      & \textbf{56.45}                   & \textbf{54.09}                   \\ 
\textsc{Gemma2-2B}              & 53.50         & 45.31             & 51.54                                                      & 44.52                                                             & 34.88                                                      & 40.85     & 56.67                                                            & 24.00      & 45.69                   & 49.41                   \\ \midrule
\multicolumn{11}{c}{\textit{Open-source models (7-14B)}} \\
\cdashline{1-11}\noalign{\vskip 0.3ex}
\textsc{LLaMA3.1-8B}            & 52.84         & 49.11             & 69.96                                                      & 67.24                                                             & 39.88                                                      & 43.80     & 59.27                                                            & 16.59      & 56.03                   & 50.98                   \\ 
\textsc{Qwen2.5-7B}             & 67.17         & 62.92             & 71.88                                                      & 81.08                                                             & 45.88                                                      & 49.83     & 77.71                                                            & 32.76      & 65.28                   & 65.05                   \\ 
\textsc{Gemma2-9B}              & 57.92         & 65.62             & 69.96                                                      & 81.28                                                             & 41.50                                                      & 49.23     & 79.17                                                            & \textbf{36.48}      & 64.23                   & 61.77 \\
\textsc{Mistral-Nemo}           & 54.25         & 55.73             & 57.38                                                      & 76.52                                                             & 41.75                                                      & 44.88     & 60.00                                                            & 33.65      & 56.11                   & 54.99                   \\ 
\textsc{Qwen2.5-14B}            & 67.50         & 70.10             & 72.83                                                      & 88.68                                                             & 53.50                                                      & 51.52     & 79.48                                                            & 31.31      & \textbf{69.20}                   & \textbf{68.80}                   \\ 
\textsc{Aya-expanse-8B}            & 65.50         & 62.40             & 44.63                                                      & 61.16                                                             & 36.88                                                      & 43.95     & 58.75                                                            & 32.77      & 49.08                   & 63.95                   \\ \midrule
\multicolumn{11}{c}{\textit{Open-source models (14-50B)}} \\
\cdashline{1-11}\noalign{\vskip 0.3ex}
\textsc{Qwen2.5-32B}            & 68.33         & 76.38             & 75.88                                                      & 90.88                                                             & 57.38                                                      & 52.27     & 83.33                                                            & 32.13      & \textbf{71.95}                   & 72.36                   \\ 
\textsc{Gemma2-27B}             & 68.00         & 64.12             & 76.67                                                      & 85.28                                                             & 50.50                                                      & 49.42     & 81.35                                                            & \textbf{42.23}      & 68.64                   & 66.06                   \\ 
\textsc{Aya-expanse-32B}            & 70.25         & 75.70             & 56.38                                                      & 86.40                                                             & 53.75                                                      & 48.33     & 64.27                                                            & 34.11      & 61.83                   & \textbf{72.98}                   \\ \midrule
\multicolumn{11}{c}{\textit{Open-source models (\textgreater{}50B)}} \\
\cdashline{1-11}\noalign{\vskip 0.3ex}
\textsc{LLaMA3.1-70B}           & 63.17         & 67.25             & 74.75                                                      & 88.28                                                             & 52.38                                                      & 55.52     & 79.17                                                            & 16.63      & 70.02                   & 65.21                   \\ 
\textsc{Qwen2.5-72B}            & 71.42         & 75.95             & 76.00                                                      & 91.00                                                             & 58.38                                                      & 52.67     & 87.60                                                            & 41.55      & \textbf{73.13}                   & \textbf{73.69}                   \\ 
\textsc{Mistral-Large}          & 69.58         & 69.04             & 77.17                                                      & 90.48                                                             & 53.50                                                      & 51.85     & 83.23                                                            & \textbf{43.40}      & 71.25                   & 69.31                   \\ \midrule
\multicolumn{11}{c}{\textit{Closed-source models}} \\
\cdashline{1-11}\noalign{\vskip 0.3ex}
\textsc{GPT-4o}                 & 69.17         & 81.04             & 77.05                                                      & 91.60                                                             & 56.75                                                      & 55.77     & 85.21                                                            & 46.32      & 73.28                   & 75.11                   \\ 
\textsc{Claude-3.7-sonnet}             & 76.13         & 81.67             & 88.58                                                      & 93.55                                                             & 67.13                                                      & 59.00    & 79.17                                                            & \textbf{48.18}      & \textbf{77.49}                   & \textbf{78.90}                 \\ \bottomrule
\end{tabular}
}

\caption{Evaluation results of different models on \textsc{P-MMEval}. We gather those models by referring to their sizes. AVG\_U and AVG\_S represent the average score of the understanding and capability-specialized tasks, respectively. \textsc{HumanEval-XL} score presents the average score of three programming languages.}
\label{tab:main_results}
\end{table*}

\subsection{\mytextcolor{Instruction selection}}
\mytextcolor{We utilize English instructions from \textsc{OpenCompass} \cite{opencompass} and \textsc{LM-Evaluation-Harness} \cite{eval-harness}. Among multiple instructions, we select a suitable one and make uniform modifications to ensure consistency across similar tasks. For zero-shot prompts, to increase the success rate of answer extraction, we add a constraint at the end of the instruction to some tasks, requiring the model to output the generated answers in a fixed format. In addition, we translate English instructions into multiple languages to construct native instructions.}

\section{\mytextcolor{Experiments}}
\mytextcolor{This section focuses on the following aspects: assessing the multilingual capabilities of different models; examining the influence of various prompts on multilingual performance; and comparing model performance in different languages.}

\subsection{Multilingual Models}
We evaluate the performance of several representative instruction-tuned models – (i) closed-source models \textsc{GPT-4o}\footnote{\texttt{gpt-4o-2024-05-13}} \cite{gpt4} and \textsc{Claude-3.7-sonnet}\footnote{\texttt{claude-3-7-sonnet-20250219}}, (ii) open-source models including \textsc{LLaMA3.1, LLaMA3.2} \cite{llama3}, \textsc{Qwen2.5} \cite{qwen2}, \textsc{Mistral-Nemo}, \textsc{Mistral-Large}, \textsc{Gemma2}, and \textsc{Aya Expanse} series \cite{aya}. 

\subsection{Evaluation Settings}
\label{sec:evaluation_settings}
According to \citet{prompts}, the choice of prompts significantly impacts the evaluation results of LLMs and the model performance is sensitive to minor variations in prompting. In this study, we compare the evaluation results using the following prompts. EN: Instructions in English + input in the target language. Native: Instructions in the target language + input in the target language. EN-Few-Shot: Instructions in English + demonstrations in the target language + input in the target language.

\mytextcolor{For MGSM, we employ Chain of Thought (CoT) \cite{cot} reasoning, which guides the model to think step-by-step before providing a final answer. For the other datasets, direct answering is utilized, which requests the model to produce answers directly. The inference methods for these datasets align with the most commonly used settings. 
Notably, for MMMLU, we choose the prompt template following OpenAI \texttt{simple-evals} repository.\footnote{\href{https://github.com/openai/simple-evals}{https://github.com/openai/simple-evals}}
Specifically, CoT reasoning exhibits a significantly higher answer extraction failure rate compared to direct answering on small-sized LLMs (i.e., the number of parameters is less than 7B), leading to poor performance. 
Thus, we employ a direct answering prompt for small-sized LLMs.\footnote{The detailed evaluation prompts are illustrated in Appendix \ref{sec:prompt_analysis}.}}

\mytextcolor{For the few-shot demonstrations, we primarily sample demonstrations from the validation set. For the missing multilingual portions, we utilize \textsc{GPT-4o} to translate these demonstrations from English into the missing languages.}

\subsection{Main Results}
Table \ref{tab:main_results} presents an overview of the evaluation results. Unless otherwise noted, the standard EN prompt is applied to all datasets except \textsc{Flores-200}, \textsc{HumanEval-XL}, and \textsc{MIFEval}, where the Native prompt is required. The evaluation result on \textsc{HumanEval-XL} is the average score across three programming languages including Python, JavaScript, and Java. See Appendix \ref{sec:performance_humaneval} for programming language evaluation details. For the Flores-200 dataset, in addition to reporting BLEU scores, we also provide COMET scores measured by wmt22-comet-da \cite{comet} (see Appendix, Table \ref{tab:comet_score}). 

First, the multilingual capabilities of models become stronger as the model sizes increase~\cite{scaling_laws}. One exception is that when the size of \textsc{LLaMA3.2} increases from 1B to 3B, there is a slight decline in performance. The main reason for this is that \textsc{LLaMA3.2-1B} and \textsc{LLaMA3.2-3B} exhibit poor instruction-following capabilities, leading to a higher failure rate in answer extraction and, consequently, fluctuations in the final score. As the model size increases, the improvements in various multilingual tasks show significant differences. Evaluation results on the understanding and capability-specialized tasks show significant improvement in understanding context, processing semantic information, reasoning, and special abilities, with increasing model sizes. For example, for the \textsc{Qwen2.5} series, the scores on the MGSM dataset for the 0.5B and 72B models are 13.12 and 91.00, respectively. In contrast, the models' performance on generation tasks is relatively weaker and shows \mytextcolor{slight} improvement. Evaluations on the \textsc{Flores-200} datasets indicate that, despite the increase in model size, the generation capability does not improve proportionally. This may reflect the complexity of generating text that maintains logical coherence and contextual relevance, where increasing model sizes does not significantly enhance output quality. 

In addition, \textsc{Qwen2.5} demonstrates a strong multilingual performance on understanding and capability-specialized tasks, while \textsc{Gemma2} excels in generation tasks. Closed-source models \textsc{GPT-4o} and \textsc{Claude-3.7-sonnet} generally outperform open-source models. The average performance gap between the best-performing open-source model and \textsc{Claude-3.7-sonnet} reaches as high as 5.21\%.

\begin{table}[!ht]
\centering
{\setlength{\tabcolsep}{5mm}
\scalebox{0.6}{%
\begin{tabular}{cccc}
\toprule
Dataset                                                     & Native & EN    & EN-Few-shot \\
\midrule
MMMLU                                                       & 44.30  & 44.69 & 45.70       \\
\textsc{MLogiQA}                                                      & 42.27  & 41.96 & 44.88       \\
MGSM                                                        & 62.13  & 64.17 & 63.28       \\
\textsc{MHellaSwag}                                                  & 52.03  & 53.37 & 59.07       \\
XNLI                                                        & 54.49  & 55.31 & 64.08       \\

\textsc{Flores-200}  & 30.00  & 24.31 & 29.18       \\ \bottomrule
\end{tabular}}
}
\caption{Comparison on \textsc{P-MMEval} using three different prompt settings. }
\label{tab:prompts_analysis}
\end{table}

\subsection{The Impact of Different Prompts on Model Performance}
\label{prompt_strategies}
We explore three different prompting strategies: EN, Native, and En-Few-Shot. Table \ref{tab:prompts_analysis} illustrates the average performance of all evaluated open-source models on various datasets of \textsc{P-MMEval}. Overall, the performance difference between the EN prompt and the Native prompt is minimal, remaining within 2\%, indicating no substantial performance gap. However, in the case of the \textsc{Flores-200}, the EN prompt results in a marked decline in performance compared to the Native prompt. We observe that models always generate responses in English when English instructions are used to describe the task for non-English data for generation tasks. On various datasets, the few-shot prompt leads to better model performance than the zero-shot prompt, as models achieve a higher success rate in extracting answers in the few-shot setting. 

\begin{figure*}[!htbp]
\centering
\includegraphics[scale=0.7]{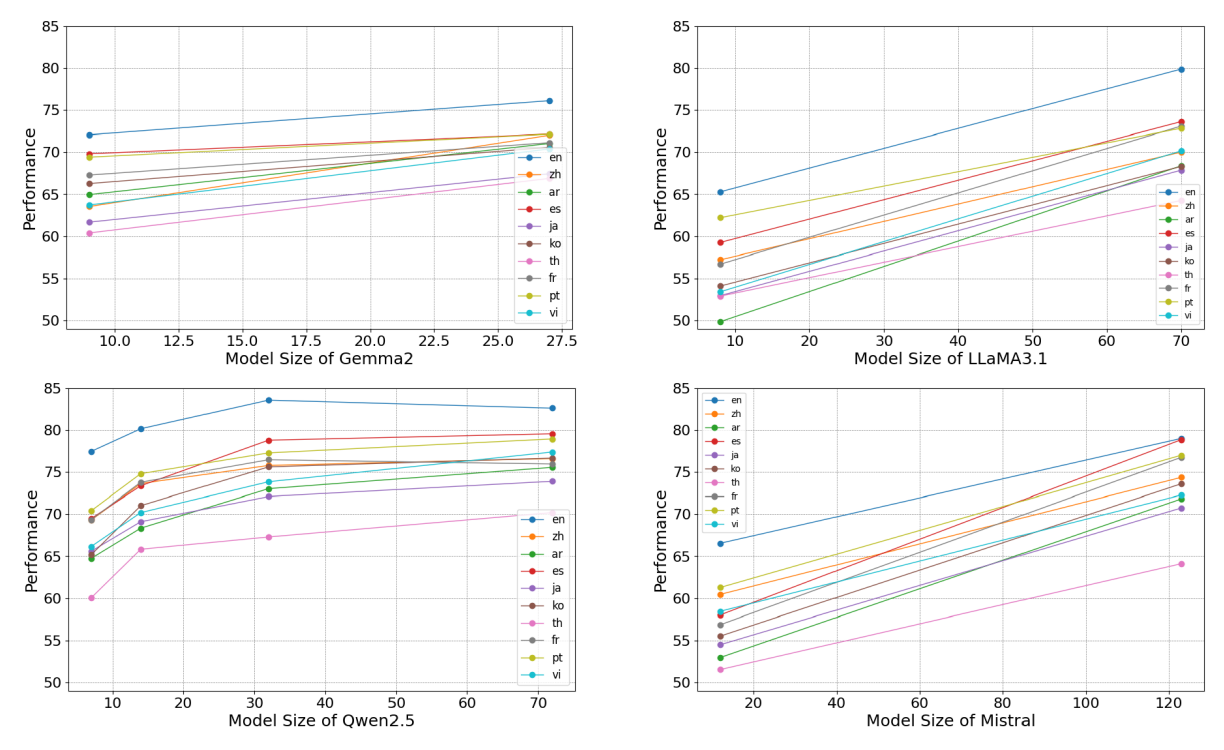}
\caption{This figure illustrates model performance for each language.}
\label{performance_across_langs}
\end{figure*}

\subsection{Language-Specific Model Performance Trends with Scale}
\label{sec:performance_across_languages}
We report the average performance per language on P-MMEval across various model sizes, excluding \textsc{MMMLU}, which is selected by models of different sizes, and \textsc{Flores-200}, which excludes English performance. In addition, we do not consider models smaller than 7B, as their performance is often highly variable and sensitive to prompt phrasing.

As shown in Fig. \ref{performance_across_langs}, model performance varies by language, with English demonstrating the strongest capabilities, followed by Spanish and Portuguese. Thai has the poorest performance, followed by Japanese. Model performance in Thai is notably inferior to other languages, with a performance gap of up to 6.64\% compared to Japanese. The distribution of training data and similarity between languages may explain these phenomena. Spanish and Portuguese are not only highly similar to English, but also have abundant language resources, reducing learning difficulty. In contrast, the Thai language has limited data resources, and Japanese belongs to an isolated language family. Another interesting phenomenon is that in Arabic, which is relatively resource-rich, smaller models around 7B perform nearly at the bottom, but when the model size increases to 14B and above, the performance significantly improves to a mid-to-low level. This indicates that Arabic has a complex linguistic system that requires more parameters to capture its complexity and nuances. 

Furthermore, we observe that in the Qwen series models (where Chinese data in the pre-training dataset is second only to English), the performance in Chinese is only mid-range, lagging behind Spanish and Portuguese. To investigate this apparent discrepancy, Appendix Section \ref{performance_comparison_w/o_zh} provides a detailed comparison of multilingual capabilities assessed on benchmarks originating from English versus those from Chinese sources. This comparative analysis reveals that the same underlying multilingual ability of a model can yield disparate evaluation outcomes and exhibit different performance distributions when assessed using benchmarks derived from different source languages.  These findings underscore a significant benchmark sensitivity in evaluating multilingual performance, indicating that the ``nativeness" or origin of the benchmark dramatically affects the observed multilingual evaluation results.

\section{Analysis of Cross-Language Transfer from English to Other Languages}
To quantitatively evaluate the model's cross-lingual transfer success rate from English to target languages, we introduce the cross-lingual accuracy consistency ratio (CACR), computed over parallel multilingual test sets. This metric assesses the proportion of instances correctly predicted in English that are also correctly predicted in the target language. The metric is formally defined in Formula \ref{cacr}, where $D_{en}$ and $D_{tgt}$ denote aligned English and target language datasets, $f(\cdot)$ represents the model's prediction function. 

\begin{figure*}
\begin{equation}
CACR_{en->tgt}=\frac{\{x|x\in D_{en}\cap D_{tgt},f(x_{en})=y_{true}\wedge f(x_{tgt})=y_{true}\}}{\{x|x\in D_{en},f(x_{en})=y_{true}\}},
\label{cacr}
\end{equation} 
\end{figure*}

\subsection{Language-Specific Transfer Capabilities and the Influence of Benchmark Origin}
We first examine the transfer success to various target languages based on benchmarks originating from English, and then compare these findings with results from a benchmark originating from Chinese to understand the impact of the benchmark's source language.

\subsubsection{Transfer Performance on English-Origin Benchmarks}
In Fig. \ref{tranfer_across_langs}, we report the average CACR for each language across the five tasks originating from English (\textsc{MGSM}, \textsc{MMMLU}, \textsc{HumanEval-XL}, \textsc{MHellaSwag}, and \textsc{XNLI}). We exclude the \textsc{Flores-200} and \textsc{IFEval} datasets, as they are not suitable for transfer analysis.

For all models, their CACR across all target languages also tends to improve as model size increases. This indicates that larger models typically possess stronger semantic representation learning and transfer capabilities. 

In addition, the difficulty of transfer varies significantly across different target languages, with Romance languages like Spanish and Portuguese showing better transfer from English, while languages like Arabic present greater challenges. Linguistic characteristics (such as lexical and syntactic similarity to English) and the coverage of the language in pre-training data are among the factors that likely influence transfer effectiveness. These performance disparities also highlight the need for more targeted optimization and data augmentation for languages with low transfer success rates.

\subsubsection{Impact of Benchmark Origin: English-Origin vs. Chinese-Origin Benchmarks}
\label{comparison_benchmark_origin}

To investigate the influence of the original language of the benchmark on perceived transfer success, we compare the results from Fig. \ref{tranfer_across_langs} (English-origin benchmarks) with those from Fig. \ref{tranfer_across_Chinese}, which reports the CACR from English for each language on a task originating from Chinese (\textsc{MLogiQA}).

When the benchmark originates from Chinese (Fig. \ref{tranfer_across_Chinese}), the CACR for transferring from English to Chinese is exceptionally high, often surpassing all other languages. In contrast, on English-origin benchmarks (Fig. \ref{tranfer_across_langs}), the CACR for Chinese, while respectable, is not as dominant. The impact of benchmark origin extends beyond just the Chinese language, leading to notable performance shifts for other languages as well. For instance, Portuguese, which demonstrates one of the highest CACR on English-origin benchmarks, sees its CACR drop to a mid-to-lower tier when the benchmark originates from Chinese. These indicate that the origin of the benchmark also affects the observed transfer success.

\subsection{Comparison of the Difficulty of Transfer in Different Tasks}
In Fig. \ref{tranfer_across_tasks}, we report the average CACR for each task across all the nine languages included in \textsc{P-MMEval}. We exclude the \textsc{Flores-200} and \textsc{IFEval} datasets.

\textbf{Model Scale Effect:}  For the \textsc{Gemma2}, \textsc{LLaMA3.1}, and \textsc{Mistral} model series, the CACR generally shows an upward trend across all six evaluated tasks as model size (parameter count) increases. However, the \textsc{Qwen2.5} model exhibits some differences. \textsc{Qwen2.5} may achieve optimal transfer performance on certain tasks at a specific scale, with larger models potentially not yielding continued benefits or even encountering optimization bottlenecks.

\textbf{Inter-task comparison:} \textsc{HumanEval-XL} (related to code generation/understanding) typically exhibits the highest CACR across all four models and various sizes. \textsc{MGSM} (mathematical reasoning) and \textsc{MMMLU} (knowledge understanding) are also consistently in the higher-performing tier, closely following \textsc{HumanEval-XL}. The transfer performance of \textsc{XNLI} (natural language inference) is typically at an upper-mid level. \textsc{MHellaSwag} (commonsense reasoning) generally performs at a lower-mid level. \textsc{MLogiQA} (logical reasoning) is almost always at the lowest performance level across all models and sizes, indicating that this type of logical reasoning capability is the most challenging for cross-lingual transfer. This ranking of task difficulty shows high consistency across different model series.

Overall, increasing model size generally enhances the average cross-lingual transfer success rate, but this is not consistently effective for all models and all tasks, with \textsc{Qwen2.5} showing transfer saturation on certain tasks. There are significant differences in the difficulty of cross-lingual transfer across tasks: code understanding and generation, mathematical reasoning, and knowledge understanding are relatively easier to transfer, while logical reasoning is the most challenging. This task difficulty hierarchy is largely consistent across different model series.

\section{Conclusion}
In this paper, we introduce a comprehensive multilingual multitask benchmark, \textsc{P-MMEval}, which covers both fundamental and capability-specialized tasks, ensuring consistent language coverage and providing parallel samples in multiple languages.
Furthermore, we conduct extensive experiments on representative multilingual model series.
These findings provide valuable guidance for future research, highlighting the importance of balanced and comprehensive training data, effective prompt engineering, and the need for targeted improvements in specific language capabilities.

\section*{Limitations}
\label{Limitations}
Through the above experiments and analyses, we summarize the following limitations: 

1) Language Coverage: While P-MMEval currently covers 10 languages from 7 language families, there is a need to include more languages to better represent global linguistic diversity. Future work will focus on expanding the language coverage to ensure a more comprehensive evaluation of multilingual LLMs. 

2) Task Diversity: P-MMEval includes eight representative tasks, but the rapidly evolving field of LLMs demands a broader range of tasks. Future work will focus on expanding the benchmark to cover more diverse and challenging tasks, providing a more thorough assessment of multilingual LLMs.

\section*{Ethics Statement}
\label{Ethics Statement}
All procedures performed in studies involving human participants were in accordance with the ethical standards of the institutional and/or national research committee and with the 1964 Helsinki Declaration and its later amendments or comparable ethical standards. This article does not contain any studies with animals performed by any of the authors. Informed consent was obtained from all individual participants included in the study.

\bibliography{main}

\appendix

\section{Sampling Process for Each Dataset in \textsc{P-MMEval}}
\label{data_sample_process}
Specifically, since \textsc{Flores}-200 already includes data for 10 languages, no additional translation was required. We retain the complete test set for evaluation.

For \textsc{HumanEval-XL} and \textsc{MGSM}, which contain 80 and 250 examples per language respectively, we ensured comprehensive coverage by translating the entire set for each language.

For single-task datasets \textsc{XNLI}, \textsc{MHellaSwag}, and \textsc{MLogiQA}, with large available test data, we follow established practices and select the first $N$ examples for translation. This approach aligns with prior literature \cite{mgsm} and ensures consistency while managing computational and resource constraints.

For multi-task datasets such as \textsc{MMMLU} and \textsc{IFEval}, we adopt different strategies. For \textsc{MMMLU}, we sample a subset comprising 200 ``hard'' samples and 200 ``easy'' samples, by utilizing diverse model evaluation results as a proxy. For \textsc{IFEval}, we select 10 examples per task type, resulting in a total of 110 examples. During the translation verification process, 14 examples were removed due to quality issues, leaving a final set of 96 examples.

\begin{table}[!ht]
\centering
\scalebox{0.8}{%
\begin{tabular}{lcc}
\toprule
Model         & COMET  & BLEU \\ \midrule
LLaMA3.2-1B   & 81.16 & 29.30 \\
LLaMA3.2-3B   & 80.58  & 36.85 \\
Qwen2.5-0.5B  & 80.06 & 15.95 \\
Qwen2.5-1.5B  & 85.17 & 21.37 \\
Qwen2.5-3B    & 87.08 & 25.75 \\
Gemma2-2B     & 86.45 & 24.00 \\ \midrule
              &        &                 \\ \midrule
LLaMA3.1-8B   & 87.16 & 16.59 \\
Qwen2.5-7B    & 87.62 & 32.76 \\
Gemma2-9B     & 88.40 & 36.48 \\
Mistral-Nemo  & 87.75 & 33.65 \\
Qwen2.5-14B   & 87.26 & 31.31 \\
Aya-expanse-8B   & 87.42 & 32.77 \\ \midrule
              &        &                 \\ \midrule
Qwen2.5-32B   & 88.56 & 32.13 \\
Gemma2-27B    & 88.83 & 42.23 \\
Aya-expanse-32B    & 88.61 & 34.11 \\ \midrule
              &        &                 \\ \midrule
LLaMA3.1-70B  & 88.27 & 16.63 \\
Qwen2.5-72B   & 88.88 & 41.55 \\
Mistral-Large & 88.76 & 43.40 \\ \bottomrule
\end{tabular}}

\caption{The table displays the comparison between BLEU and COMET scores on the Flores-200 dataset.}
\label{tab:comet_score}
\end{table}

\section{Evaluation of COMET Scores on the Flores-200 Dataset}
\label{sec:comet_score}
In addition to the BLEU scores, we also provide COMET scores measured using the wmt22-comet-da model, shown in Table \ref{tab:comet_score}. For all tested models, the COMET scores are significantly higher than the BLEU scores, indicating that COMET is a more forgiving evaluation metric. Unlike BLEU, which requires strict literal matching, COMET focuses more on the semantics and fluency of the translation. 

Additionally, COMET scores for all tested models are consistently high, generally ranging between 80 and 90, with negligible score differences observed between some models of large size gaps. This clustering of high scores and minimal variation indicates that COMET, in this specific evaluation scenario, likely lacked sufficient discriminative power to effectively measure nuanced performance differences between the various models or sizes. Consequently, we opt not to use COMET and continue to rely on BLEU as the primary evaluation metric for translation results, which, despite its own limitations, could still offer some relative performance insights in this context.

\begin{table}[!ht]
\centering
\scalebox{0.8}{%
\begin{tabular}{cccc}
\toprule
              & Python & JavaScript & Java  \\ \midrule
\textsc{LLaMA3.2-1B}   & 92.13  & 9.38       & 11.63 \\ 
\textsc{LLaMA3.2-3B}   & 91.50  & 9.75       & 11.00 \\ 
\textsc{Qwen2.5-0.5B}  & 78.38  & 14.25      & 9.13  \\ 
\textsc{Qwen2.5-1.5B}  & 81.63  & 35.88      & 28.25 \\ 
\textsc{Qwen2.5-3B}    & 84.00  & 53.75      & 44.50 \\ 
\textsc{Gemma2-2B}     & 98.13  & 29.25      & 27.25 \\ \midrule
              &        &            &       \\ \midrule
\textsc{LLaMA3.1-8B}   & 96.38  & 46.88      & 66.63 \\ 
\textsc{Qwen2.5-7B}    & 86.75  & 68.00      & 60.88 \\ 
\textsc{Gemma2-9B}     & 98.75  & 54.63      & 56.50 \\ 

\textsc{Mistral-Nemo}  & 93.25  & 39.63      & 39.25 \\ 
\textsc{Qwen2.5-14B}   & 84.50  & 72.75      & 61.25 \\ 
\textsc{Aya-expanse-8B}     & 72.63  & 30.13      & 31.13 \\ \midrule
              &        &            &       \\ \midrule
\textsc{Qwen2.5-32B}   & 89.38  & 73.13      & 65.13 \\ 
\textsc{Gemma2-27B}    & 99.63  & 63.75      & 66.63 \\ 
\textsc{Aya-expanse-32B}     & 96.25  & 39.00      & 33.88 \\ \midrule
              &        &            &       \\ \midrule
\textsc{LLaMA3.1-70B}  & 98.75  & 63.38      & 62.13 \\ 
\textsc{Qwen2.5-72B}   & 85.63  & 75.00      & 67.38 \\ 
\textsc{Mistral-Large} & 88.63  & 73.88      & 69.00 \\ \midrule
              &        &            &       \\ \midrule
\textsc{GPT-4o}        & 89.13  & 77.88      & 64.13 \\ 
\textsc{Claude-3.7-sonnet}    & 98.38  & 81.50      & 88.58 \\ \bottomrule
\end{tabular}}

\caption{The table presents the performance on three programming languages of HumanEval-XL.}
\label{tab:humaneval-xl}
\end{table}

\section{Evaluation Results on Three Programming Languages of HumanEval-XL}
\label{sec:performance_humaneval}
Table \ref{tab:humaneval-xl} shows the evaluation results of all tested models on three programming languages of HumanEval-XL. Model performance in Python greatly exceeds the performance in the other two programming languages. For instance, Gemma2-2B scores 98.13 in Python, compared to 29.25 in JavaScript and 27.25 in Java. Additionally, as the model size increases, there is a noticeable improvement in performance for both JavaScript and Java.

\section{Comparison of the Multilingual Performance on Tasks originating from English and Chinese}
\label{performance_comparison_w/o_zh}
\begin{figure*}[!htbp]
\centering
\includegraphics[scale=0.7]{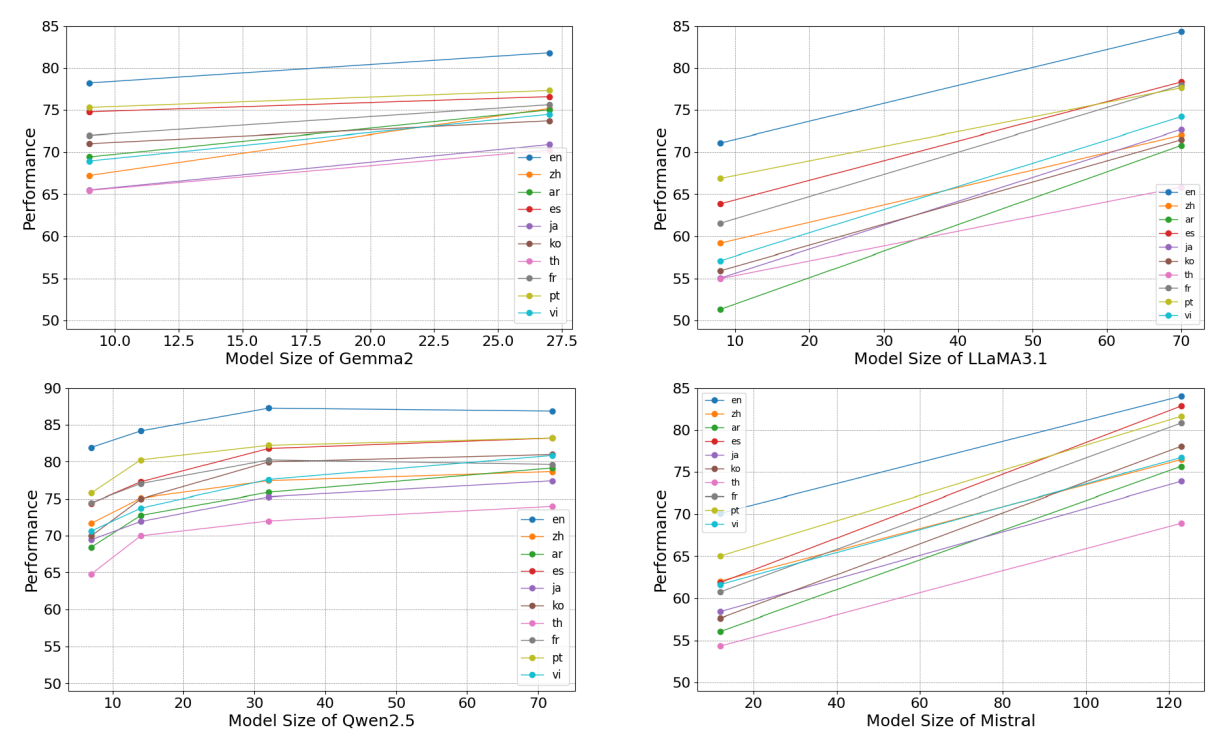}
\caption{This figure presents the average performance for each language on English-sourced tasks.}
\label{performance_across_langs_without_zh}
\end{figure*}

\begin{figure*}[!htbp]
\centering
\includegraphics[scale=0.7]{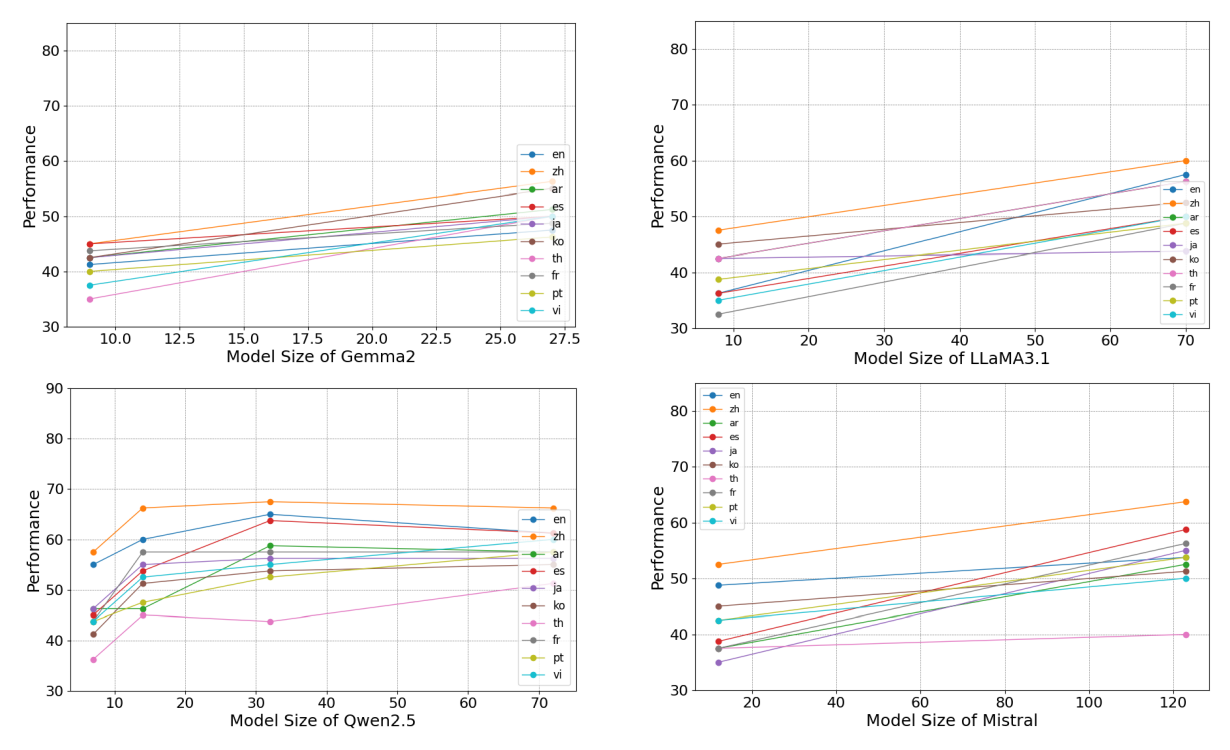}
\caption{This figure shows the average performance for each language on Chinese-sourced tasks.}
\label{performance_across_langs_zh}
\end{figure*}

On English-sourced benchmarks (Fig. \ref{performance_across_langs_without_zh}), the model performs best in English, followed by excellent performance in Spanish and Portuguese (fellow Indo-European languages), and only moderate performance in Chinese. Conversely, on Chinese-sourced benchmarks (Fig. \ref{performance_across_langs_zh}), the model performs best in Chinese, followed by English, while Portuguese performance is only mediocre. Notably, Japanese performance is among the lowest on English-sourced benchmarks, surpassing only Thai. However, performance improves to a mediocre level on Chinese-sourced benchmarks. This difference may be due to lexical similarities between Japanese and Chinese. We suggest that when benchmarks are translated into other languages, the translation process itself, or inherent linguistic and cultural nuances, might inadvertently increase the difficulty for languages that are structurally and culturally more distant from the native languages.

\section{Analysis of Cross-Language Transfer from English to Other Languages}
\begin{figure*}[!htbp]
\centering
\includegraphics[scale=0.7]{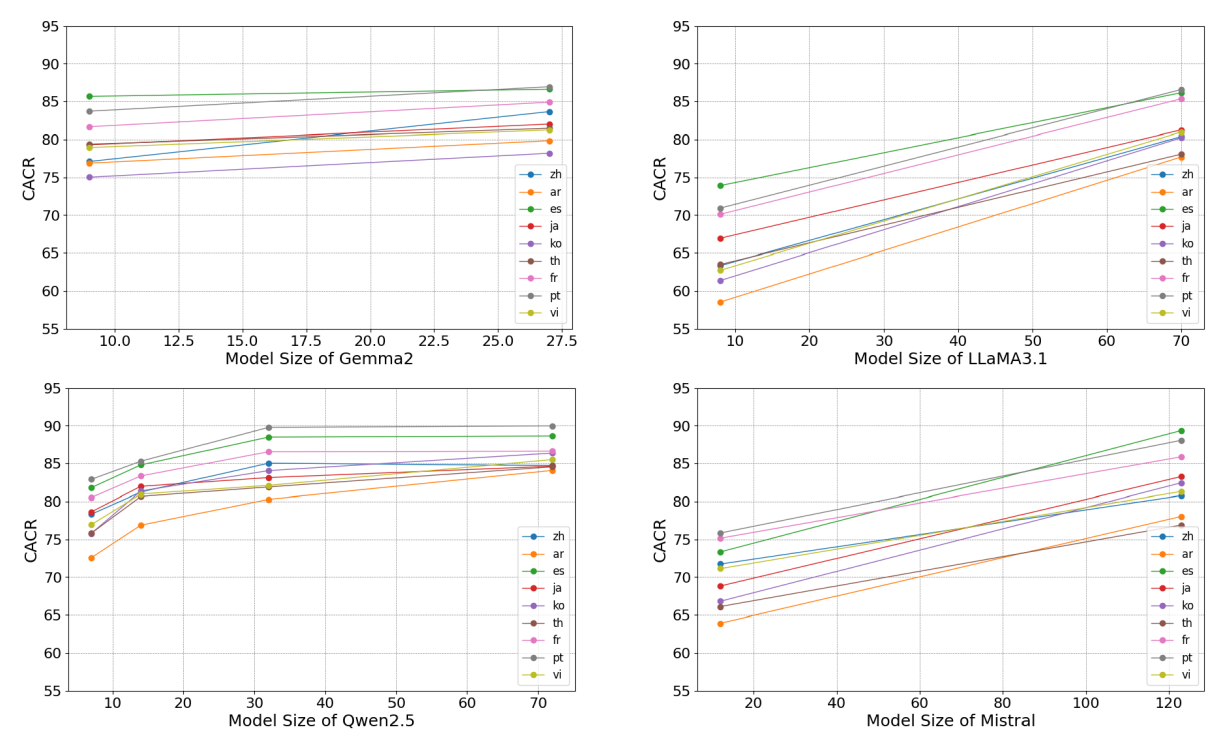}
\caption{This figure illustrates the average CACR for each language on English-sourced tasks.}
\label{tranfer_across_langs}
\end{figure*}

\begin{figure*}[!htbp]
\centering
\includegraphics[scale=0.7]{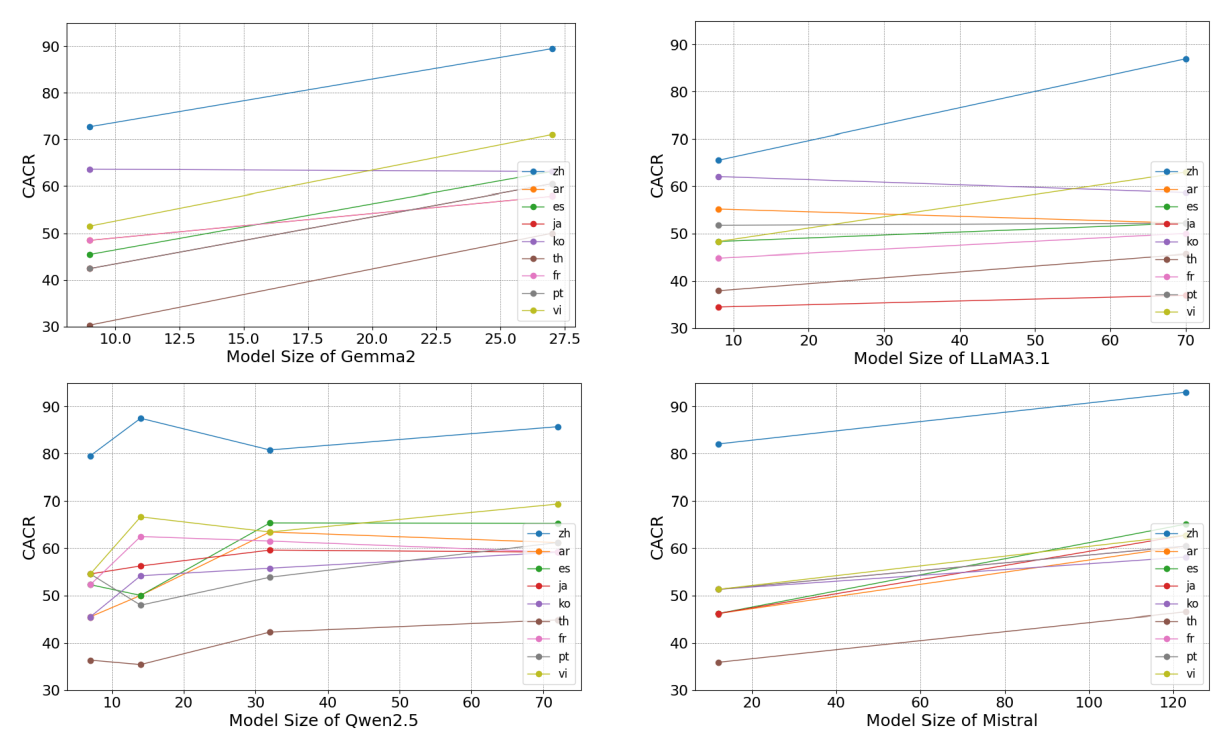}
\caption{This figure illustrates the average CACR for each language on the MLogiQA task originating from Chinese.}
\label{tranfer_across_Chinese}
\end{figure*}

\begin{figure*}[!htbp]
\centering
\includegraphics[scale=0.7]{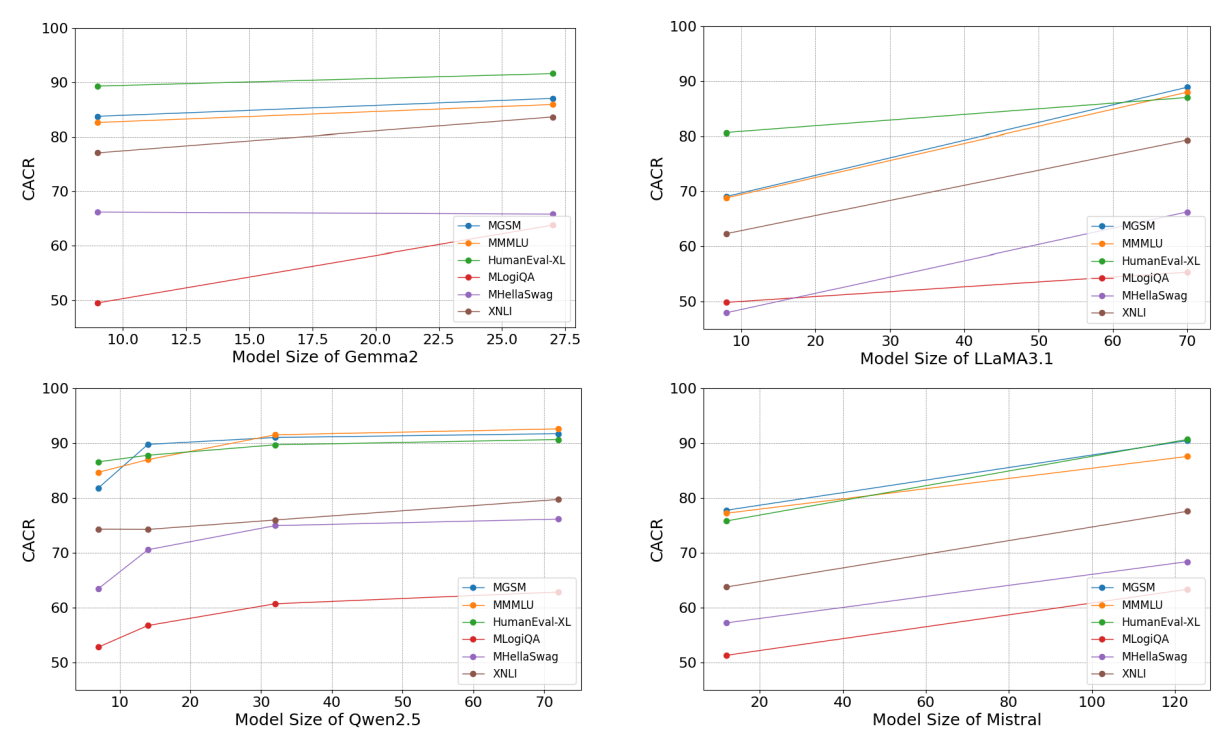}
\caption{This figure displays the average CACR transferring from English to all target languages, broken down by task.}
\label{tranfer_across_tasks}
\end{figure*}

Fig. \ref{tranfer_across_langs} and Fig. \ref{tranfer_across_Chinese} illustrate the average CACR for each language on tasks originating from English and Chinese, respectively. Fig. \ref{tranfer_across_tasks} illustrates the CACR for each task. 

\section{The Prompt Utilized for Each Dataset}
\label{sec:prompt_analysis}
The section presents the inference prompt utilized for each dataset. 

\begin{figure*}[!htbp]
\centering
\includegraphics[scale=0.45]{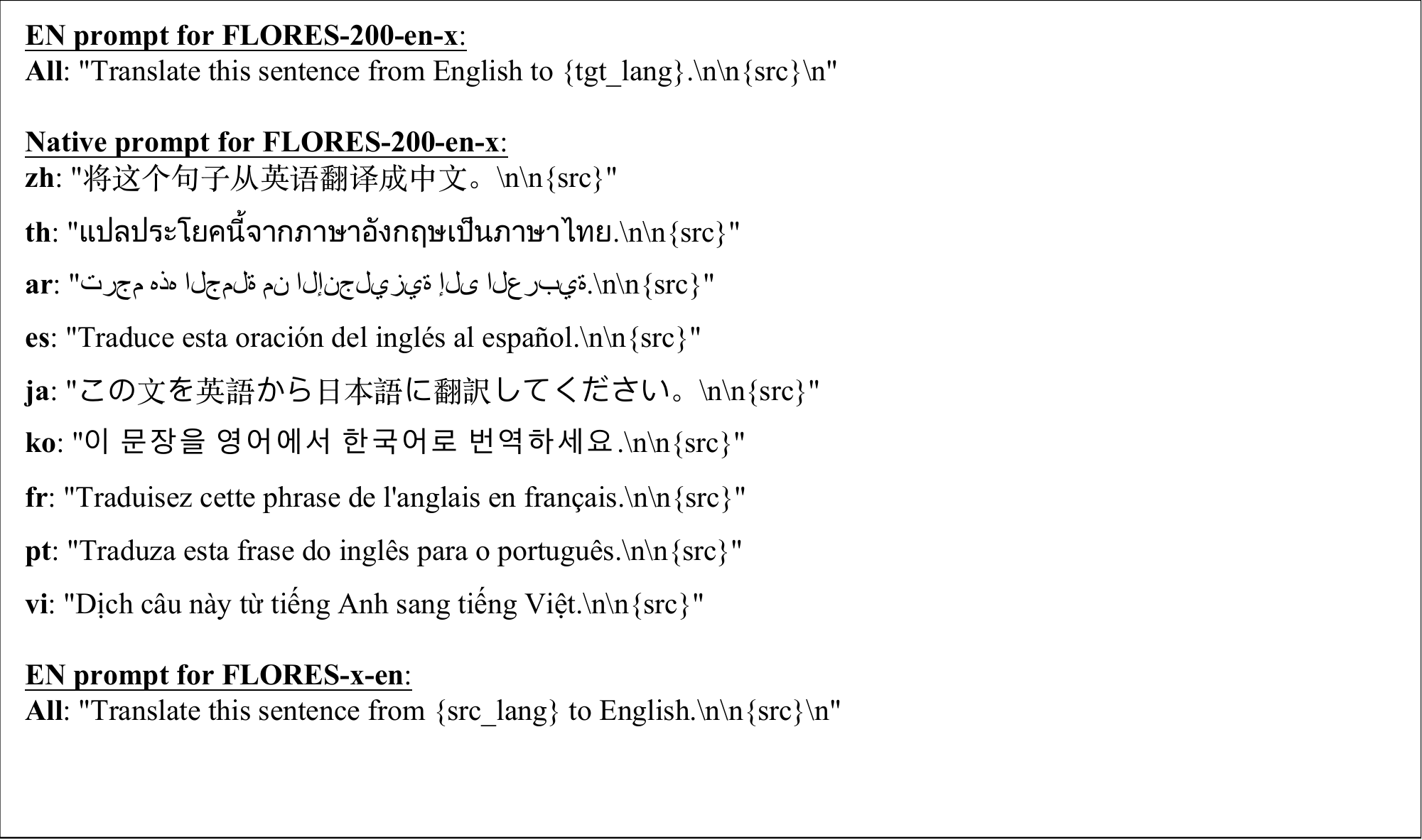}
\caption{This figure presents the prompt for the Flores-200 dataset.}
\label{flores_prompt}
\end{figure*}

\begin{figure*}[!htbp]
\centering
\includegraphics[scale=0.45]{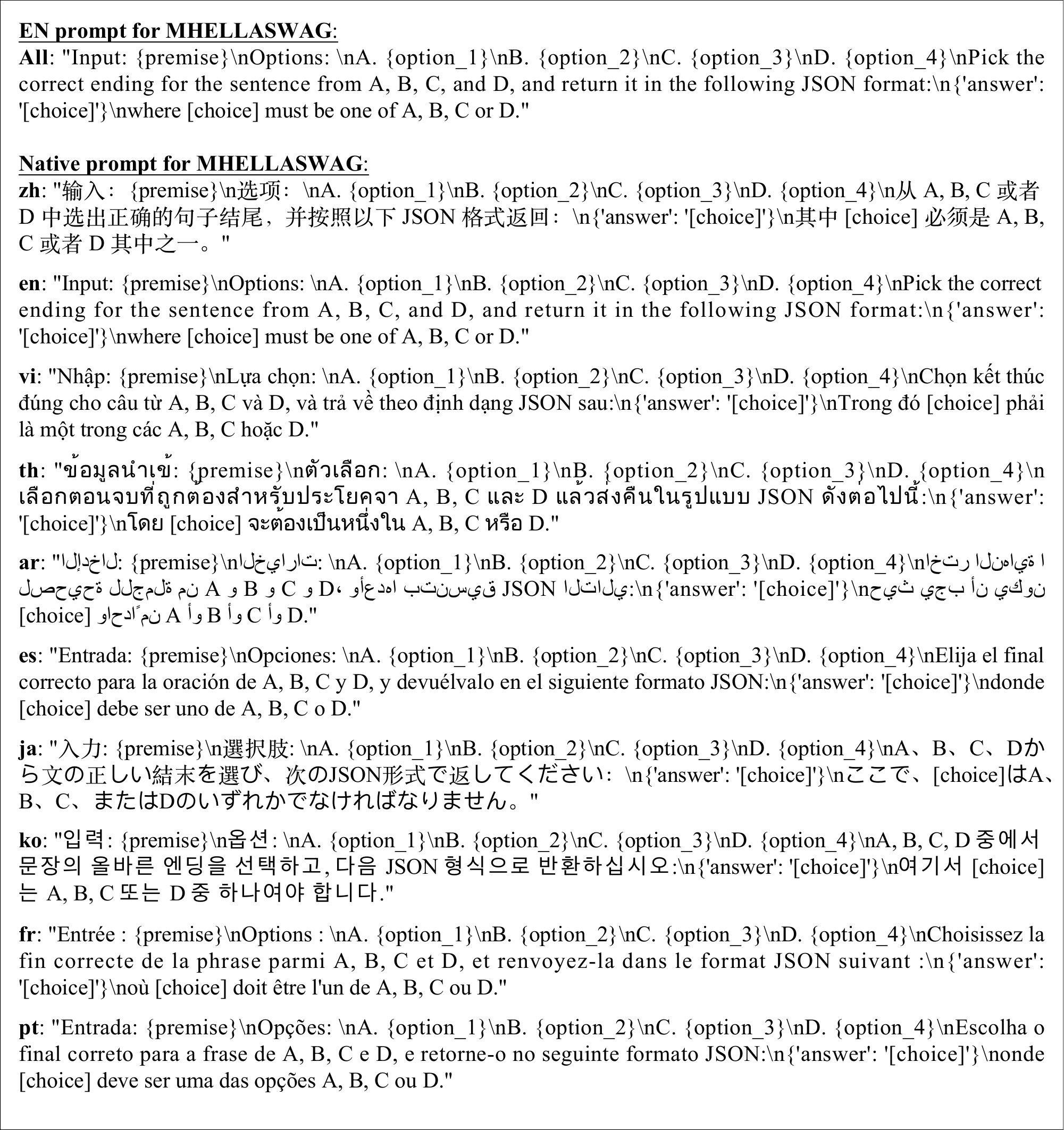}
\caption{This figure presents the prompt for the MHellaSwag dataset.}
\label{mhellaswag_prompt}
\end{figure*}

\begin{figure*}[!htbp]
\centering
\includegraphics[scale=0.45]{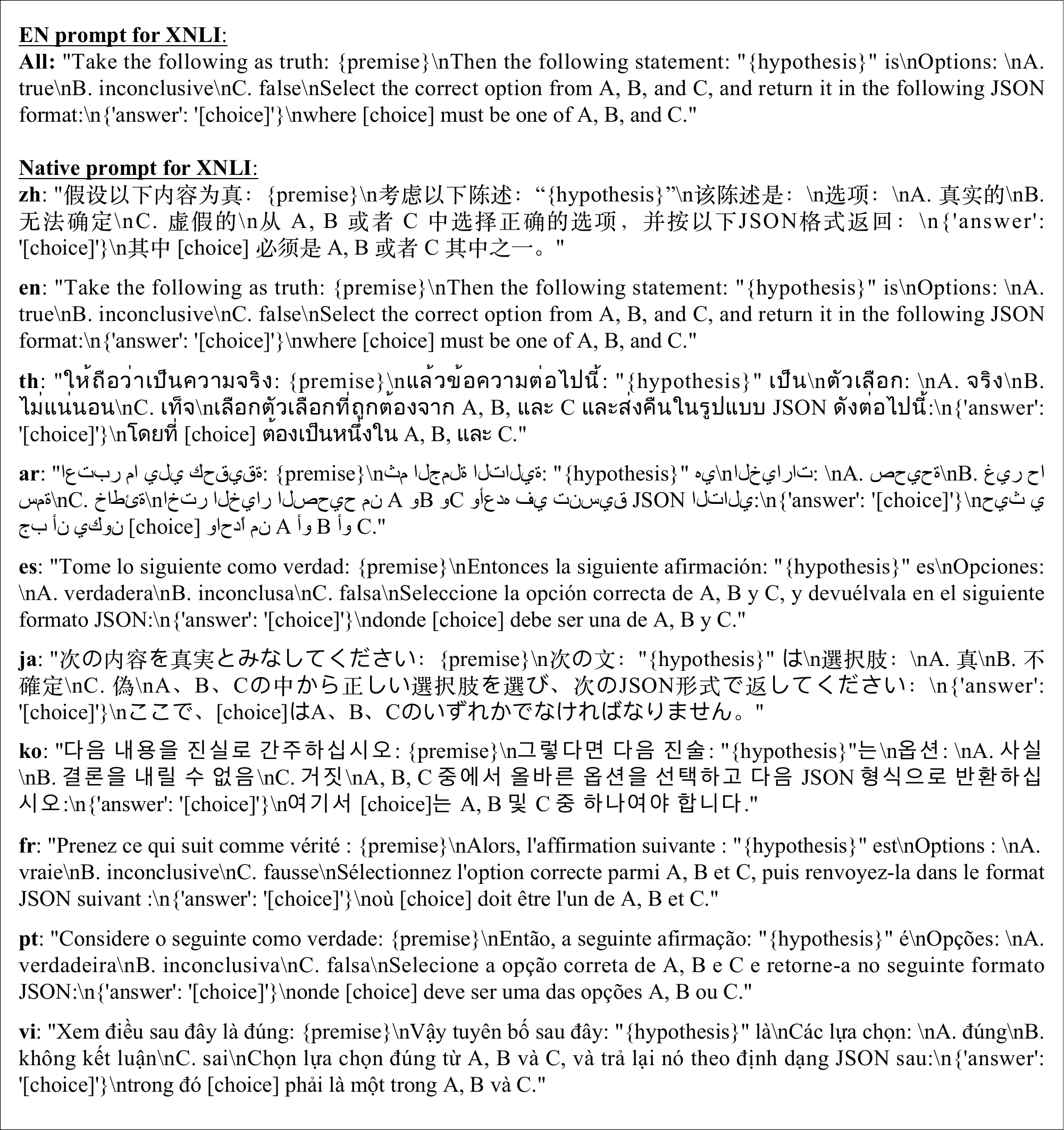}
\caption{This figure presents the prompt for the XNLI dataset.}
\label{xnli_prompt}
\end{figure*}

\begin{figure*}[!htbp]
\centering
\includegraphics[scale=0.45]{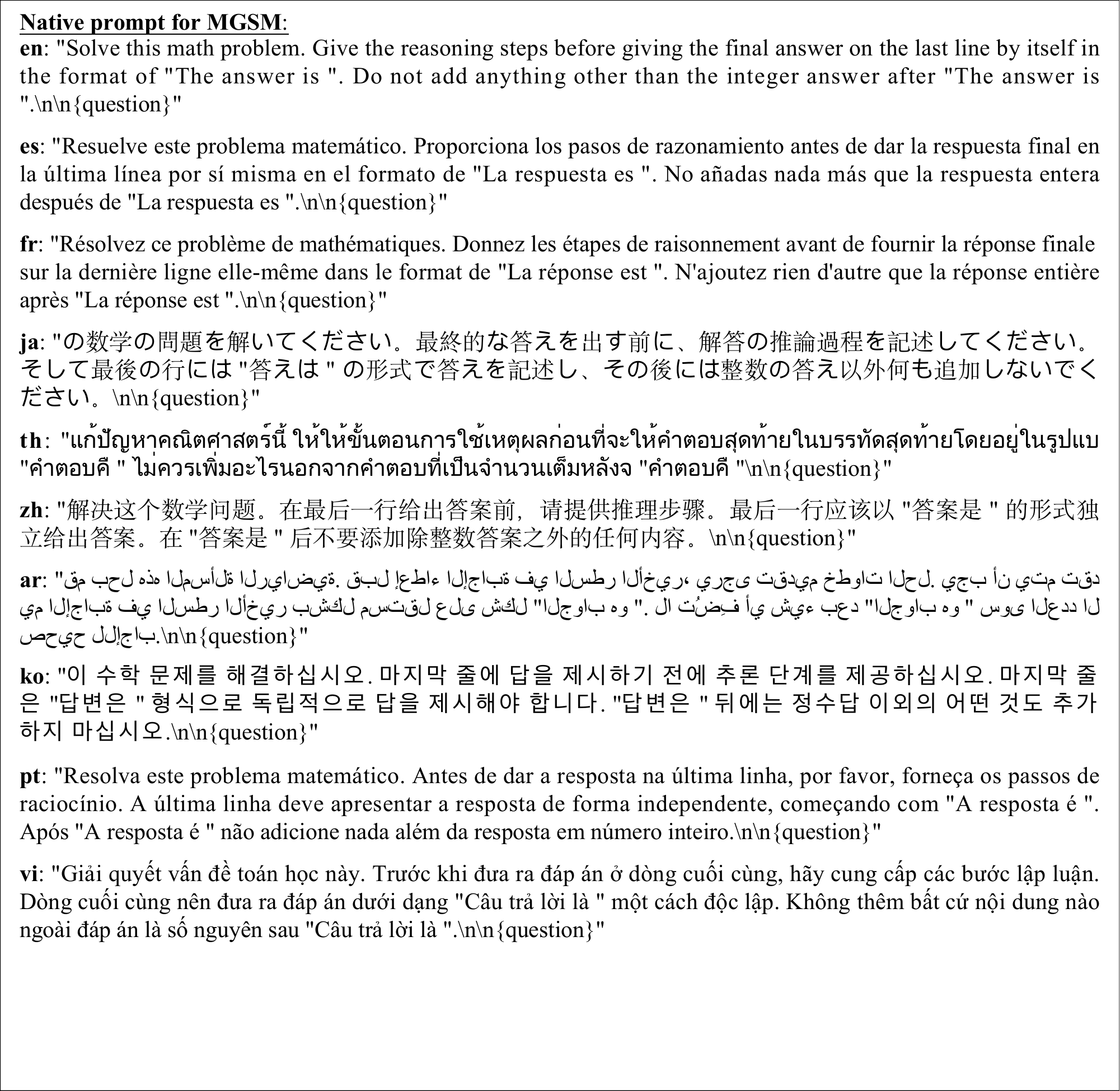}
\caption{This figure presents the Native prompt for the MGSM dataset.}
\label{mgsm_native_prompt}
\end{figure*}

\begin{figure*}[!htbp]
\centering
\includegraphics[scale=0.45]{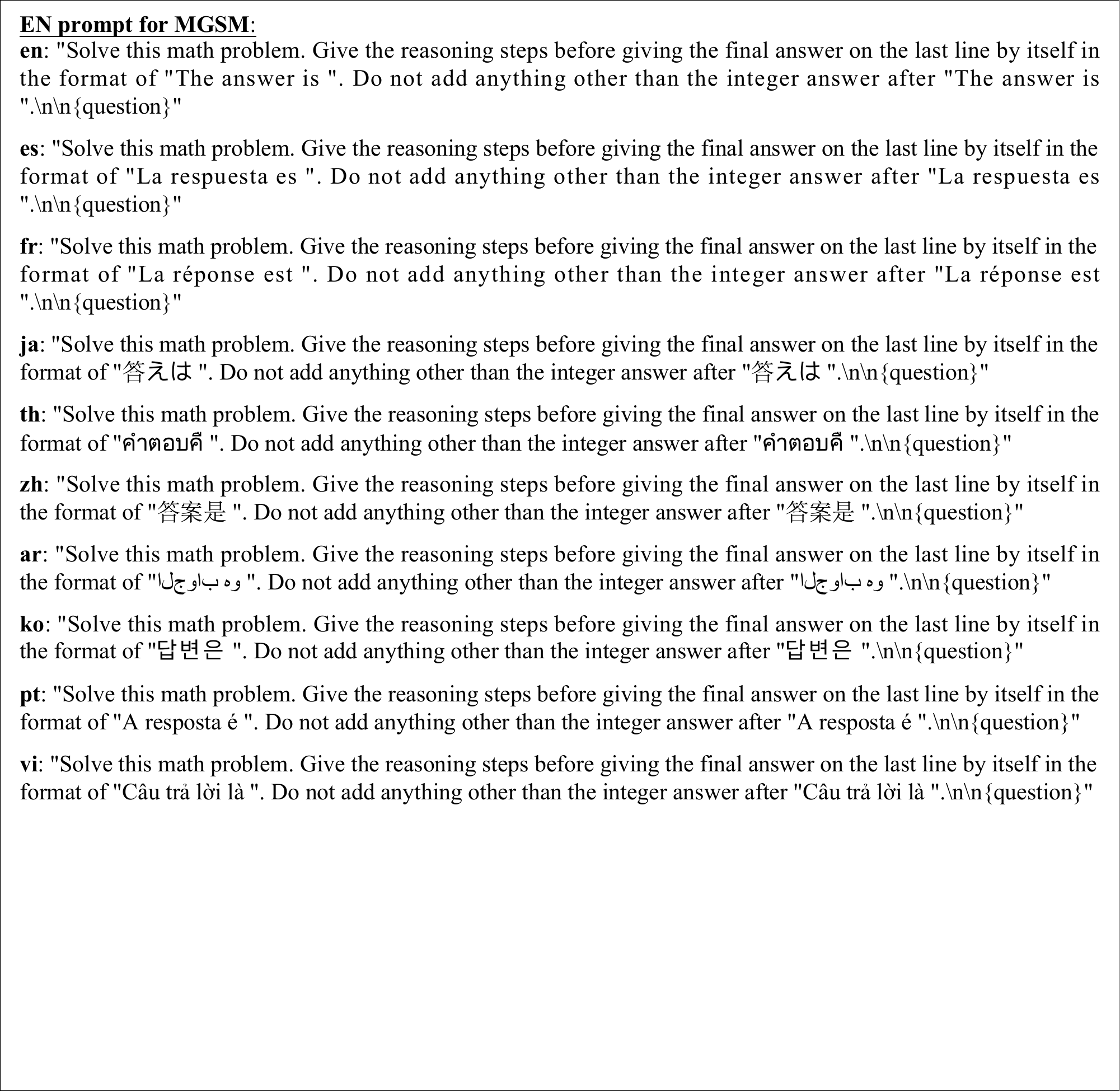}
\caption{This figure presents the EN prompt for the MGSM dataset.}
\label{mgsm_en_prompt}
\end{figure*}

\begin{figure*}[!htbp]
\centering
\includegraphics[scale=0.45]{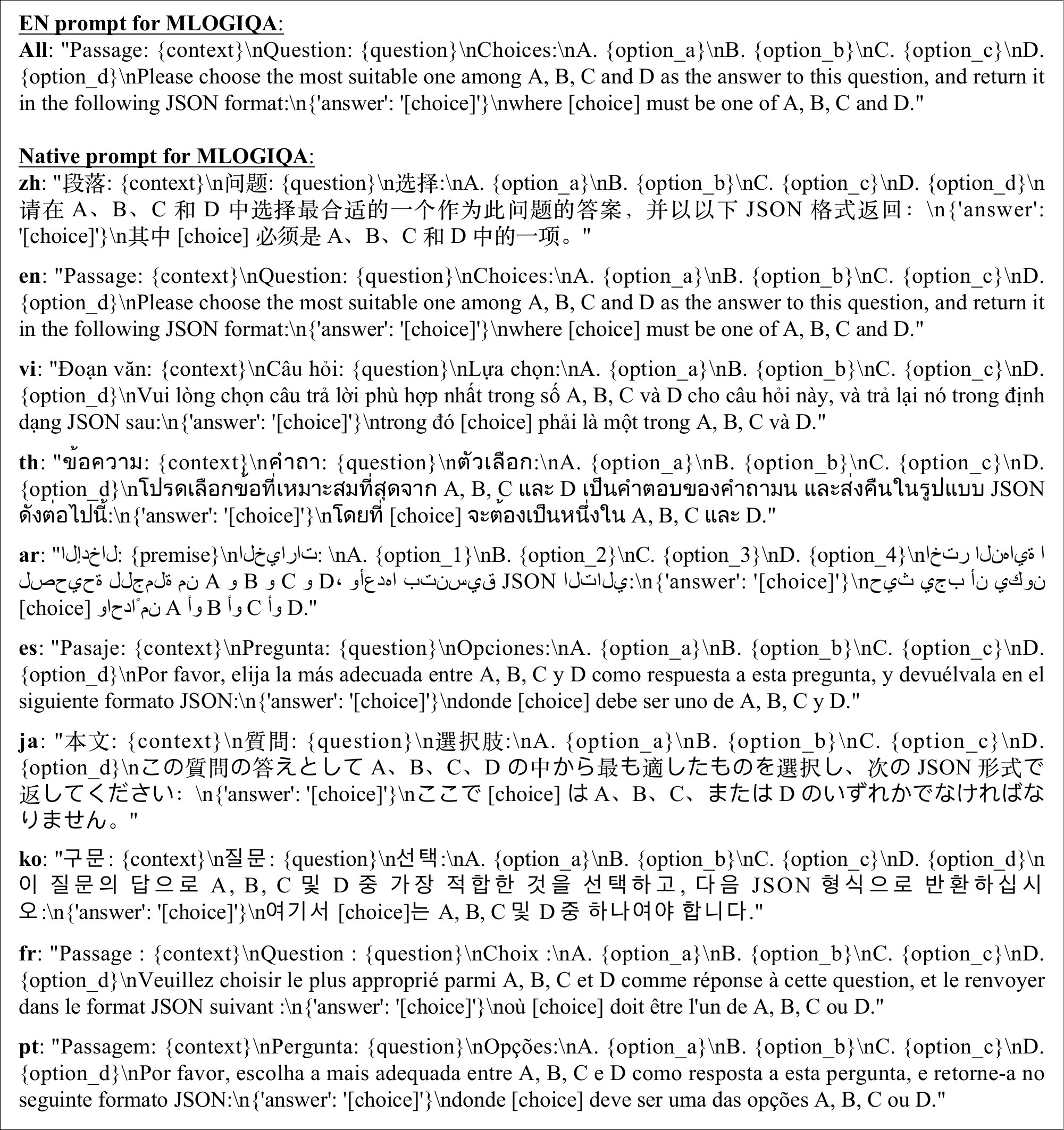}
\caption{This figure presents the prompt for the MLogiQA dataset.}
\label{logiqa_prompt}
\end{figure*}

\begin{figure*}[!htbp]
\centering
\includegraphics[scale=0.45]{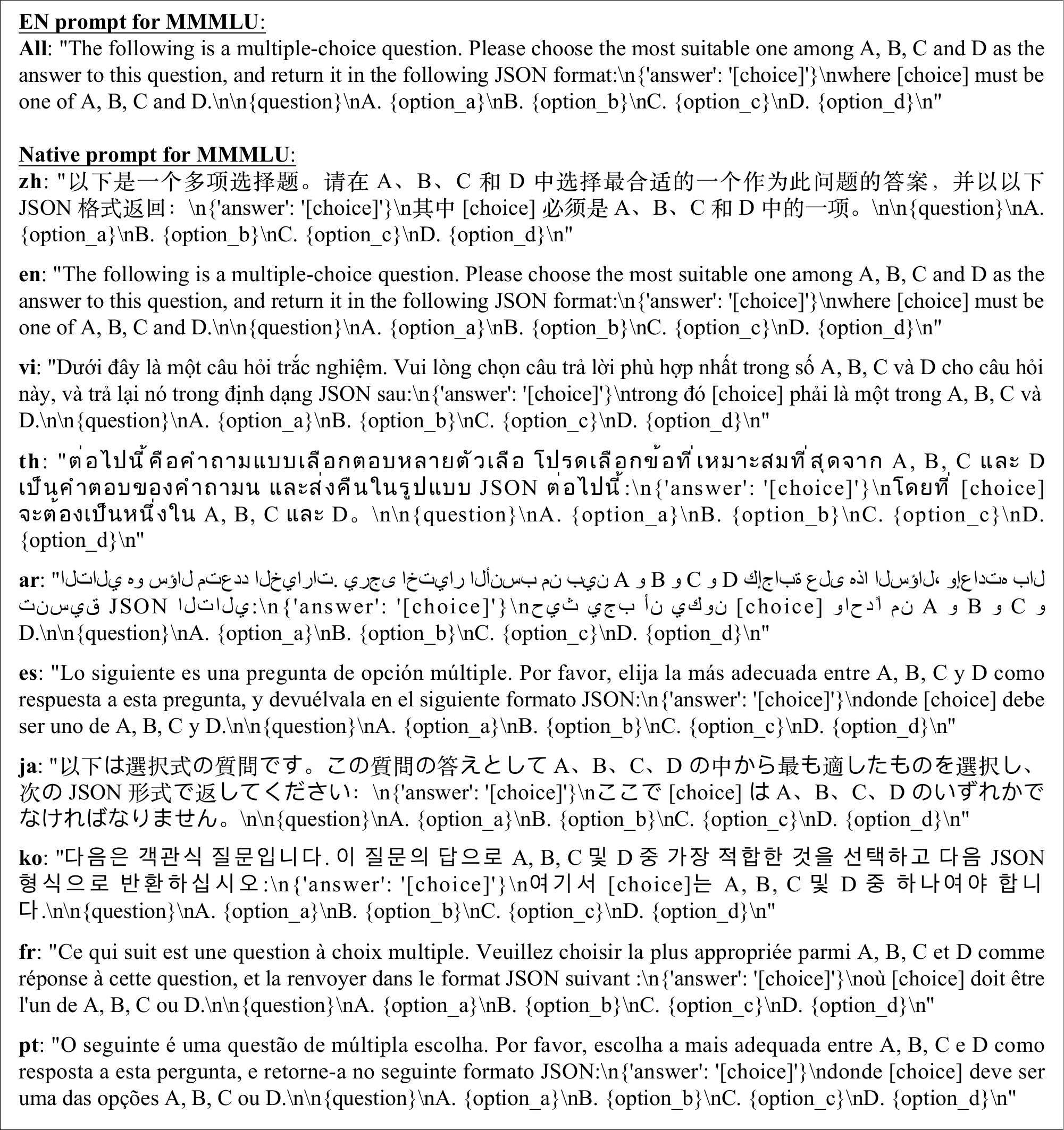}
\caption{This figure presents the prompt for the MMMLU dataset.}
\label{mmmlu_prompt}
\end{figure*}

\end{document}